\documentclass[letterpaper]{article} % DO NOT CHANGE THIS
\usepackage{aaai2027}  % DO NOT CHANGE THIS
\usepackage[hyphens]{url}  % DO NOT CHANGE THIS
\usepackage{graphicx} % DO NOT CHANGE THIS
\usepackage{natbib}  % DO NOT CHANGE THIS AND DO NOT ADD ANY OPTIONS TO IT
\usepackage{caption} % DO NOT CHANGE THIS AND DO NOT ADD ANY OPTIONS TO IT
\usepackage{algorithm}
\usepackage{algorithmic}

\usepackage{newfloat}
\usepackage{listings}
\DeclareCaptionStyle{ruled}{labelfont=normalfont,labelsep=colon,strut=off} % DO NOT CHANGE THIS
\floatstyle{ruled}
\newfloat{listing}{tb}{lst}{}
\floatname{listing}{Listing}

\usepackage{amsmath,amsfonts}
\usepackage{booktabs}
\usepackage[table,xcdraw]{xcolor}
\usepackage{multirow}
\usepackage{makecell}
\usepackage{tabularx}
\usepackage{array}
\newcolumntype{Y}{>{\centering\arraybackslash}X}
\nocopyright

\title{From Synthetic to Real: Toward Identity-Consistent Makeup Transfer with Synthetic and Real Data}
\author{
    Yue Yu\textsuperscript{\rm 1},
    Jiayu Wang\textsuperscript{\rm 1},
    Jiajia Shi\textsuperscript{\rm 1},
    Zhiyu Tan\textsuperscript{\rm 2},
    Hao Li\textsuperscript{\rm 2},
    Jingjing Chen\textsuperscript{\rm 1}\corresponding,
}
\affiliations{
    \textsuperscript{\rm 1}College of Computer Science and Artificial Intelligence, Fudan University\\
    \textsuperscript{\rm 2}{Shanghai Academy of Artificial Intelligence for Science
}
}

\begin{document}

\maketitle

\begin{abstract}
Makeup transfer aims to apply the makeup style of a reference portrait to a source portrait while preserving identity and background. Recent methods leverage synthetic supervision to improve performance, yet two challenges remain: (1) synthetic supervision frequently fails to faithfully preserve identity, and (2) the domain gap between synthetic and real data limits generalization, resulting in degraded performance in complex real-world scenarios. To address these issues, we first propose ConsistentBeauty, a novel data curation pipeline that ensures makeup fidelity and strict identity consistency within the synthesized data. Second, we propose RealBeauty, a synthetic-to-real post-training framework. Beyond supervised learning on curated synthetic data, we further adapt the model to real-world scenarios through reinforcement learning and design novel verifiable rewards tailored to the makeup transfer task. In addition, we establish a new diverse benchmark for makeup transfer, covering a wide range of skin tones, ages, genders, poses, and makeup styles, thereby enabling a more comprehensive evaluation of model performance under diverse real-world conditions. Extensive experiments show that our method achieves state-of-the-art performance on multiple benchmarks and demonstrates clear advantages in identity preservation and performance on complex real-world cases.
\end{abstract}


\section{Introduction}

Makeup transfer aims to seamlessly render the makeup from a given portrait image onto a source portrait. It requires both accurate makeup transfer and faithful identity preservation, while keeping background regions unchanged. Given its practical value in social media, e-commerce and the beauty industry, this task has attracted increasing attention in computer vision. However, although a non-makeup source portrait and a reference makeup image are readily available, the corresponding transferred result is generally unavailable for direct supervision. As a result, makeup transfer remains a challenging task.

\begin{figure}[t]
        \centering
        \includegraphics[width=8.3cm]{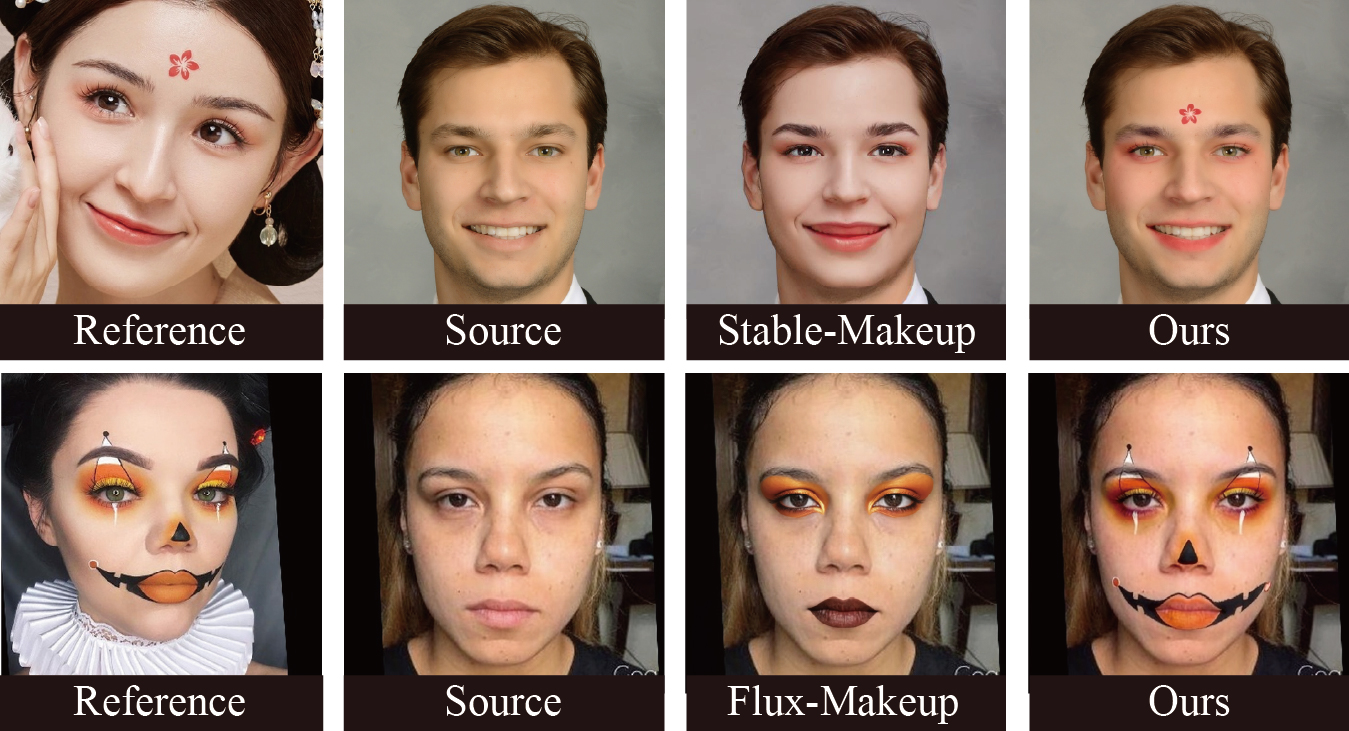}
        \caption{Existing methods usually suffer from two key limitations: {\bf Top row}: failure to preserve source identity; {\bf Bottom row}: degraded performance under complex makeup.}
        \label{limitations}
        \vspace{-5mm}
\end{figure}

Existing methods can be broadly divided into two categories. The first category treats makeup transfer as an unsupervised image-to-image translation task and commonly follows a CycleGAN-style \cite{zhu2017unpaired} training scheme. These methods \cite{li2018beautygan, jiang2020psgan, yan2023beautyrec} adopt surrogate objectives which often inadequately capture the desired makeup effects, resulting in limited transfer capability. The second category focuses on synthesizing pseudo ground-truth images to enable supervised training. Early works \cite{yang2022elegant, sun2022ssat} perform affine transformations and color matching to “paste” the reference makeup onto the source portrait, generating coarse-grained pseudo ground truth. With the rapid development of diffusion and flow-based generative models \cite{rombach2022high, esser2024scaling}, more recent studies \cite{zhang2025stablemakeup, yang2025ffhq, zhu2025flux, wu2025evomakeup} employ image editing models to synthesize high-quality makeup and non-makeup pairs for supervised training. Benefiting from this, these methods achieve substantially improved transfer capability. Nevertheless, this line of work still suffers from two limitations, as shown in Figure~\ref{limitations}. First, these methods do not always faithfully preserve the source identity, as the models used to synthesize training data may introduce unintended changes to facial characteristics. Second, although these methods work well for simple and common makeup patterns, their performance degrades in complex real-world cases due to the gap between synthetic training data and real-world scenarios.

These observations suggest that effective makeup transfer requires addressing two issues simultaneously, i.e., (1) how to construct training data that preserves both makeup fidelity and identity consistency; and (2) how to bridge the gap between synthetic training data and real-world scenarios. To this end, we first introduce {\bf ConsistentBeauty}, a novel data curation pipeline. Specifically, we collect a large number of non-makeup portraits and construct high-quality makeup layers covering light, heavy, and artistic styles. Rather than relying on image editing models, we employ a 3D triangulation-based method to seamlessly apply the same makeup layer to two non-makeup images with different identities. This process enables us to construct training triplets $\langle${\it non-makeup source, makeup reference, target image}$\rangle$  for supervised training. In this way, our data construction pipeline ensures strict identity consistency and high makeup fidelity. To further reduce the gap between synthetic training data and real-world cases, we then propose {\bf RealBeauty}, a synthetic-to-real post-training framework for makeup transfer. RealBeauty adopts a two-stage training strategy, where the model is first cold-started on our curated dataset and then further optimized via reinforcement learning on unpaired real-world makeup data. During the reinforcement learning stage, we design a new task-specific verifier that accurately measures the makeup fidelity. In this way, our post-training framework enables the model to jointly leverage synthetic supervision and real-world data, thereby improving real-world generalization.

In addition, we establish a new benchmark for the makeup transfer task. Existing studies mostly rely on earlier datasets \cite{li2018beautygan, jiang2020psgan, gu2019ladn} with randomly sampled pairs, which lack sufficient diversity. We curate a new benchmark containing 300 non-makeup portraits and 300 makeup images. The non-makeup images cover different skin tones, ages, genders, and poses, while the makeup images include diverse styles from light to heavy and from simple to complex. This benchmark enables a more comprehensive evaluation of models' transfer capability under various conditions.

To summarize, our key contributions are threefold:

\begin{itemize}
    \item We propose a data curation pipeline and construct a large-scale dataset of high visual quality, strong makeup consistency and strict identity consistency for supervised training in makeup transfer. Additionally, we establish a diverse and challenging benchmark that enables comprehensive evaluation of makeup transfer models.
    \item We introduce a synthetic-to-real post-training framework for makeup transfer and design a novel verifiable reward, enabling the model to learn jointly from synthetic and real data. Our method achieves state-of-the-art performance on multiple benchmarks.
    \item Extensive experiments demonstrate the advantages of our curated dataset and the effectiveness of the proposed method.
\end{itemize}

\section{Related Work}

\subsection{Makeup Transfer}
Makeup transfer has attracted continuous attention owing to its practical value. Early approaches were based on GAN \cite{goodfellow2020generative}. In the absence of real ground-truth as supervision, methods such as BeautyGAN \cite{li2018beautygan}, BeautyRec \cite{yan2023beautyrec}, and PSGAN \cite{jiang2020psgan} were trained in an unsupervised image-to-image translation manner. They typically adopted surrogate objectives such as histogram matching to compute makeup-related losses, thereby enforcing makeup consistency. EleGANt \cite{yang2022elegant} and SSAT \cite{sun2022ssat} constructed pseudo ground truth (PGT) using simple matching and affine transformation techniques as supervisory signals. With the paradigm shift in image generation models in recent years, more methods have adopted diffusion-based frameworks for makeup transfer. SHMT \cite{sun2024shmt} decouples makeup information from human identity features and trains the model in a self-supervised manner. Stable-Makeup \cite{zhang2025stablemakeup}, FLUX-Makeup \cite{zhu2025flux}, and EvoMakeup\cite{wu2025evomakeup} employ image editing models to synthesize paired makeup and non-makeup samples for supervised training. FLUX-Makeup and EvoMakeup are built upon flow-based models and leverage more powerful image editing models \cite{labs2025flux, wang2025seededit} for data construction. Compared with previous methods, these models exhibit stronger makeup transfer capabilities. However, the synthetic makeup supervision often fails to preserve the original source identity, leading to degraded identity consistency and limiting model performance.

% \subsection{Diffusion and Flow-based Generative Models}
% In recent years, diffusion models \cite{song2019generative, ho2020denoising, song2020score, song2020denoising,ramesh2022hierarchical} have developed rapidly, demonstrating remarkable effectiveness across a wide range of generation tasks \cite{ye2023ipadaptertextcompatibleimage,ruiz2023dreambooth,brooks2023instructpix2pix}. Compared with GAN-based image generation models \cite{goodfellow2020generative,mirza2014conditional, wang2018perceptual}, diffusion models exhibit better training stability and stronger generative capability. As defined in DDPM \cite{ho2020denoising}, diffusion models generate images by iteratively denoising a random noise, gradually transforming it into a meaningful image. Stable Diffusion \cite{rombach2022high} further advanced this paradigm by performing the noising and denoising processes in the latent space, significantly reducing computational costs. DiT \cite{peebles2023scalable} introduced a Transformer-based architecture that replaces the conventional U-Net used in previous diffusion models. Stable Diffusion 3 \cite{esser2024scaling} and Flux integrate flow mechanisms \cite{liu2022flow, lipman2022flow} and transformer-based diffusion models, further enhancing the models' image generation capabilities. 

\subsection{Reinforcement Learning for Diffusion Models}
Reinforcement learning has emerged as a key technique in the post-training phase of large language models (LLMs). When equipped with appropriate rewards, it enables LLMs to better align with specific preferences or further enhance their capabilities in particular domains. In contrast to LLMs, reinforcement learning for diffusion models \cite{song2019generative, ho2020denoising, song2020score, song2020denoising,ramesh2022hierarchical} and flow-based models \cite{liu2022flow, lipman2022flow} remains in an initial stage of development. Earlier works \cite{lee2023aligning}  directly adopted rewards as training objectives for supervised learning. Diffusion-DPO \cite{wallace2024diffusion} formalizes the DPO \cite{rafailov2023direct} loss for diffusion models, enabling direct optimization on pairwise preference data. Inspired by the GRPO algorithm introduced in DeepSeekMath \cite{shao2024deepseekmath}, subsequent works \cite{liu2025flow, xue2025dancegrpo, li2025mixgrpo} begin exploring the application of GRPO for optimizing diffusion models. In this work, by performing reinforcement learning,  the model can be trained on real data rather than solely on synthetic data.

\begin{figure*}[t]
        \centering
        \includegraphics[width=16.8cm]{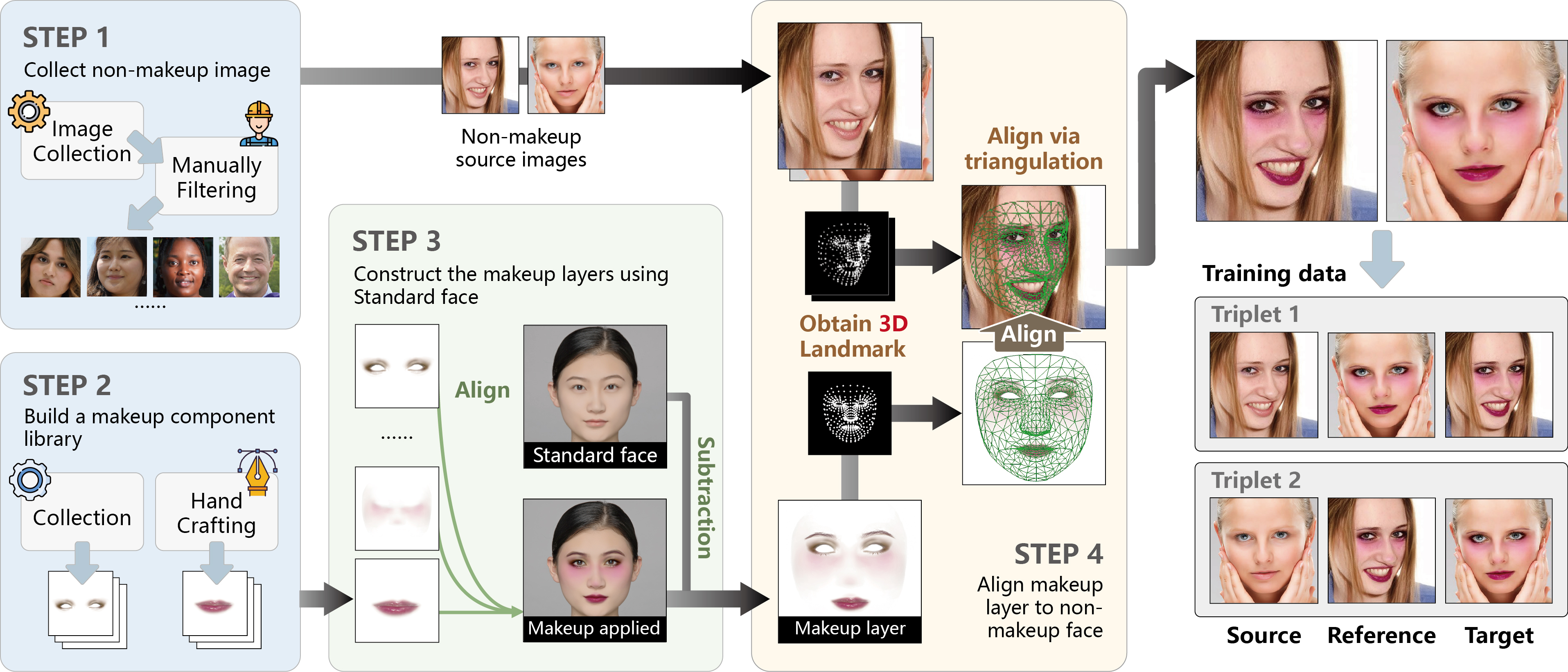}
        \caption{Overview of the proposed ConsistentBeauty data curation pipeline. By constructing makeup layers on a standard face and transferring them to other non-makeup portraits via 3D triangulation, the proposed pipeline constructs high-quality triplets with strict identity consistency and high makeup fidelity, thereby providing reliable supervision for makeup transfer.}
        \label{data_curation}
\end{figure*}

\section{Method}
In this section, we provide a detailed description of our data curation pipeline, dataset, synthetic-to-real post-training framework, and benchmark.

\subsection{Data Curation}

\subsubsection{Data Curation Pipeline}
The goal of makeup transfer is to apply the makeup style of the reference image $I_{ref}$ onto the person in the source image $I_{src}$. The generated results should faithfully reflect the reference makeup style while preserving the identity of $I_{src}$. To ensure both makeup fidelity and identity consistency in synthetic data, we propose a data curation pipeline, {\bf ConsistentBeauty}, that produces high-quality triplets $\langle I_{src}, I_{ref}, I_{tgt}\rangle$, where $I_{tgt}$ denotes the target result of the transfer, serving as a supervision signal. In each triplet, $I_{tgt}$ shares the same identity as $I_{src}$ and the same makeup style as $I_{ref}$. The curation pipeline is illustrated in Figure~\ref{data_curation} and consists of four steps as follows.

{\bf Step 1}. We first collect non-makeup face images from the Internet and the publicly available FFHQ dataset \cite{karras2019style}. These images cover diverse skin tones, genders, ages, lighting conditions, and poses. After manually filtering out samples with unfavorable poses, severe occlusions, or unsatisfactory image quality, we obtain 19,639 high-quality non-makeup portraits. 

{\bf Step 2}. We curate a high-quality makeup component library by collecting and manually crafting makeup components in seven categories: eyebrows, eyeliners, eyelashes, eyeshadows, blushes, lipsticks, and complex facial patterns. 

{\bf Step 3}. We construct the makeup layers. The collected makeup components are randomly combined and aligned to a non-makeup standard face $I^{std}_{non}$ using facial landmarks, resulting in a makeup-applied standard face $I^{std}_{make} $. The makeup layer $\mathrm{layer}_{make}$ is then defined as the pixel-wise residual between the makeup-applied and non-makeup standard faces in the color space, i.e., $\mathrm{layer}_{make} = I^{std}_{make} - I^{std}_{non}$. By leveraging the 3D facial geometry of the standard face $I^{std}$, the resulting makeup layer can be transferred to other non-makeup faces, producing more natural and seamless results.

{\bf Step 4}. Using 3D triangulation, the constructed makeup layer is aligned to non-makeup portrait $I_{src}$ in 3D space. Finally, two source images with the same makeup layer can serve as references for each other, forming two triplets of $\langle I_{src}, I_{ref}, I_{tgt}\rangle$.

% \begin{figure}[t]
%         \centering
%         \includegraphics[width=8.3cm]{images/dataset.jpg}
%         \caption{Examples from our ConsistentBeauty dataset. Different non-makeup images within the same row are applied with the same makeup style, demonstrating strict identity consistency and high makeup fidelity in the constructed triplets. In particular, the sticker-style data shown in the third row is useful for improving the model's ability to transfer complex facial patterns.}
%         \label{dataset}
%         \vspace{-3mm}
% \end{figure}

\subsubsection{Constructed Dataset}
Building upon our non-makeup portrait collection and the makeup component library, we construct a total of 290,144 triplets to form the {\bf ConsistentBeauty} dataset. Compared with existing data construction methods and datasets, our dataset offers three advantages: {\bf Strict identity consistency and high makeup fidelity}. Within each triplet, the identities in $I_{src}$ and $I_{tgt}$ are strictly consistent, while $I_{ref}$ and $I_{tgt}$ share the same makeup appearance. {\bf High scalability}. Makeup components are disentangled from one another and from identity, allowing them to be randomly combined into new makeup layers and applied to different identities. This enables efficient dataset expansion.

% {\bf Clearer Supervision.} Compared to methods that construct only image pairs, our triplet-based formulation is more suitable for makeup transfer, as it provides clearer supervision by explicitly separating the identity and makeup information, enabling more effective learning of the task.

\subsubsection{Ethics Statement}
The Institutional Review Board has approved the construction of the ConsistentBeauty dataset. The dataset is intended exclusively for academic research. Access is restricted to scholarly use and requires formal application.

\subsection{Synthetic-to-Real Post-Training Framework}
To bridge the gap between synthetic training data and real-world scenarios, we propose {\bf RealBeauty}, a two-stage post-training framework built on a pretrained flow-based generative model. We first perform supervised fine-tuning (SFT) on ConsistentBeauty to establish basic makeup transfer capability, and then apply reinforcement learning (RL) to real-world data without ground-truth supervision. To support the latter stage, we design a task-specific Makeup Perceptual Verifier that provides a reliable makeup-fidelity reward.

\begin{figure*}[t]
        \centering
        \includegraphics[width=15.5cm]{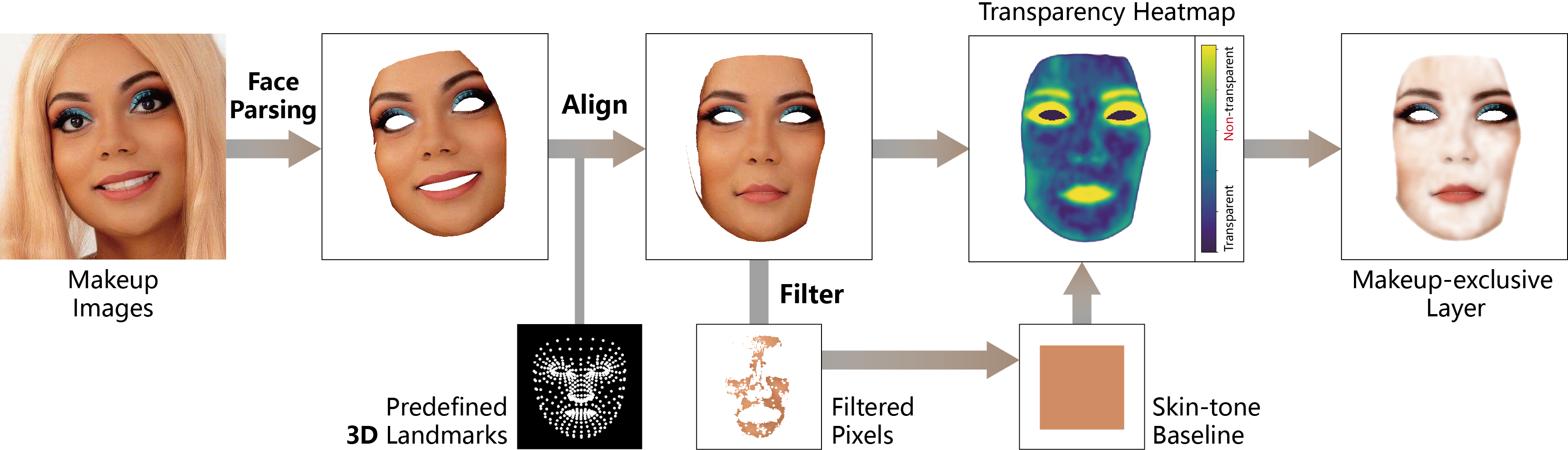}
        \caption{Overview of the proposed Makeup Perceptual Verifier for extracting a makeup-exclusive layer from the input image. By removing background regions and reducing the influence of identity and skin tone, the verifier extracts makeup-focused representations for more accurate evaluation of makeup transfer quality.}
        \label{verifier}
        \vspace{-2mm}
\end{figure*}

\subsubsection{Training Procedure}
The SFT stage serves as a cold start, where the model is optimized on ConsistentBeauty using the rectified flow-matching loss:
\begin{equation}
\begin{aligned}
\mathcal L_{SFT}= &\mathbb E_{t\sim p(t),x_{tgt},x_{src},x_{ref},\epsilon} \\
&[||\pi(z_t,t,x_{src},x_{ref})-(\epsilon-x_{tgt})||_2^2]
\end{aligned}
\end{equation}
where $\pi$ represents the model, $t$ denotes the timestep, and $x_{tgt}$, $x_{src}$ , $x_{ref}$ are the latents encoded from  $I_{tgt}$, $I_{src}$ , $I_{ref}$, respectively.  $\epsilon$ is standard Gaussian noise and $z_t=(1-t)x_{tgt}+t\epsilon$. Subsequently, we perform RL on real-world data using GRPO~\cite{shao2024deepseekmath, liu2025flow} with two rewards. For each rollout, the Makeup Perceptual Verifier provides a makeup-fidelity reward, while face similarity~\cite{schroff2015facenet} provides an identity-consistency reward. Their group-normalized advantages are denoted by $A_{\text{makeup}}$ and $A_{\text{face}}$, respectively. However, due to the inherent trade-off between makeup fidelity and identity consistency, directly summing their advantages may cause one reward to dominate the optimization. To balance the two rewards, we incorporate an additional synergy term~\cite{huang2026competition}:
\begin{equation}
    S = \mathrm{sgn}(\mathrm{min}(A_{\text{makeup}},A_{\text{face}}))\mathrm{tanh}(|A_{\text{makeup}}A_{\text{face}}|),
\end{equation}
where $\mathrm{sgn}(\cdot)$ denotes the sign function. This term is positive only when both advantages are positive. The final advantage is defined as: 
\begin{equation}
    A = \lambda_{\text{makeup}} A_{\text{makeup}} + \lambda_{\text{face}} A_{\text{face}} + \alpha S, 
\end{equation}
where $\lambda_{\text{makeup}}$, $\lambda_{\text{face}}$ and $\alpha$ are the corresponding weights. Thus, the synergy term encourages simultaneous improvements in makeup fidelity and identity consistency while penalizing degradation in either.

\subsubsection{Makeup Perceptual Verifier}
Learning from real-world data without ground-truth supervision requires a reliable reward. A natural choice is the evaluation metric, which measures whole-image similarity between the generated and reference images. However, this metric is affected by makeup-irrelevant information, including facial identity and background, resulting in inaccurate assessments. More critically, it may encourage identity drift from the source toward the reference. We therefore design a task-specific Makeup Perceptual Verifier that suppresses such irrelevant information, improving makeup fidelity without compromising identity consistency. We first apply face parsing to obtain the facial region $R_{face}$ to eliminate the influence of background. We then align it to a predefined 3D landmark template via facial triangulation to reduce the influence of identity-specific facial geometry. Finally, to suppress skin-tone information in $R_{\text{face}}$, we use a skin-tone filtering algorithm to assign transparency to non-makeup skin regions. Specifically, we use a Gaussian filter to remove high-frequency details such as eyes, lips, and decorative makeup patterns, while retaining smooth low-frequency components associated with skin tone. For the filtered pixels, we further remove the boundary chroma and luminance values in their distribution to exclude potential smooth makeup regions, such as blush. The average hue, chroma, and luminance of the remaining pixels define the skin-tone baseline. As larger deviations from this baseline indicate a higher likelihood of makeup, we compute a soft transparency value for every pixel in $R_{face}$:
\begin{equation}
    \mathrm{Transparency} = 1-e^{-(\lambda_h \Delta H + \lambda_c \Delta C + \lambda_l \Delta L)}
\end{equation}
where $\Delta H$, $\Delta C$ and $\Delta L$ denote the hue, chroma, and luminance differences from the skin-tone baseline, respectively, and $\lambda$ is their weights. Consequently, non-makeup pixels close to the skin-tone baseline become transparent, whereas makeup pixels with larger deviations remain opaque. Applying this transparency map to the reference and generated images yields their makeup-exclusive layers, as illustrated in Figure~\ref{verifier}. In reinforcement learning, we compute the cosine similarity between the DINO \cite{caron2021emerging} embeddings extracted from the makeup-exclusive layers of the reference and generated images, which serves as the verifiable reward.

\subsection{Benchmark}
% 数字要改
Existing benchmarks have two major limitations. First, their source and reference images are dominated by frontal portraits of light-skinned women, limiting evaluation across skin tones, genders, and poses. Second, their reference images mainly contain simple makeup, making them inadequate for evaluating complex makeup transfer. To enable more comprehensive evaluation, we introduce BeautyBench. Its source set contains 300 FFHQ portraits \cite{karras2019style} with diverse skin tones and a balanced gender distribution. Of these portraits, 70\% are frontal and 30\% are profile faces. Its reference set comprises 300 images collected from online sources and CPM-Real \cite{nguyen2021lipstick}, including 115 light-makeup, 115 heavy-makeup, and 70 complex face-painting images. Random pairing yields a 1,000-pair test set covering diverse facial characteristics and makeup styles. To prevent data leakage, all images in BeautyBench are strictly disjoint from the data used in both the SFT and RL stages.

\begin{figure*}[t]
        \centering
        \includegraphics[width=16.2cm]{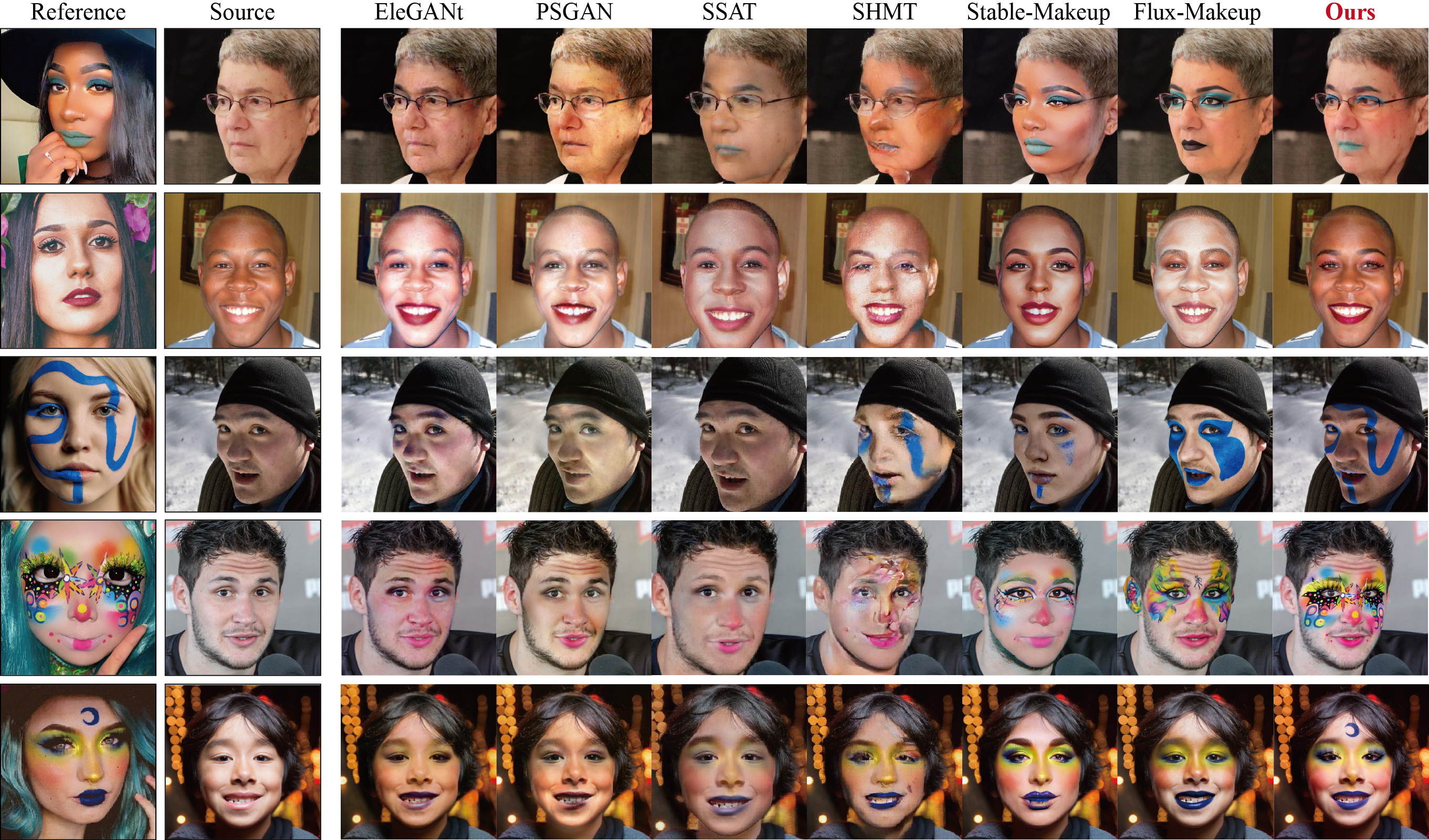}
        \caption{Qualitative comparison with representative methods. Existing methods mainly exhibit two limitations: (1) insufficient identity consistency with the source image. For example, in the first two rows, some methods show noticeable identity shift toward the reference. (2) limited transfer capability in complex cases. In the last three rows, most methods struggle to accurately reproduce complex makeup patterns. In contrast, our method achieves better identity consistency and higher makeup fidelity.}
        \label{comparison}
\end{figure*}

\begin{figure*}[t]
        \centering
        \includegraphics[width=17.8cm]{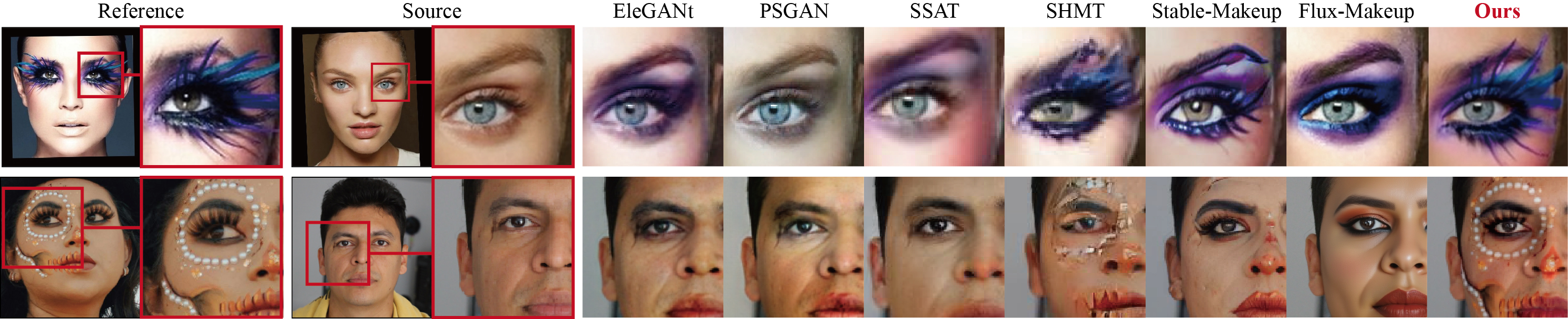}
        \caption{The zoomed-in regions highlight fine-grained details, such as decorative elements and intricate makeup patterns. The comparisons show our method generates realistic and high-quality details, which are highly consistent with the reference makeup.}
        \label{details}
\end{figure*}

\section{Experiments}

\subsection{Experimental Settings}

%{\bf Implementation Details}. We take the FLUX.1-dev as pretrained model and trained the model with a LoRA \cite{hu2022lora} of 512 rank throughout the entire experiment. During supervised fine-tuning, the model is trained on the ConsistentBeauty dataset for 20,000 steps using 4 A100 GPUs, with a batch size of 8 per GPU and a learning rate of 1e-5. In the reinforcement learning stage, we utilize real makeup images collected from the internet and combine the reward signals from makeup perceptual verifier, face similarity, and image reward \cite{xu2023imagereward} as the final reward, with equal weights assigned to each. The model undergoes 240 iterations of training on 8 A100 GPUs  with a batch size of 1 per GPU and a learning rate of 2e-6. Inference was performed with a step size of 25 and a guidance scale of 4.
\subsubsection{Implementation Details.}
% 查一下
We build upon FLUX.1-dev and optimize LoRA adapters with 512 rank. Training consists of SFT on ConsistentBeauty and RL utilizing real makeup images collected from the internet. The RL reward is based on our Makeup Perceptual Verifier and Face Similarity. Detailed settings are provided in the supplementary material.
%{\bf Evaluation Datasets}. In addition to evaluating on BeautyBench, we assess the methods on two makeup transfer datasets—MT \cite{li2018beautygan} and LADN \cite{gu2019ladn} following existing works. For each dataset, 1,000 source-reference image pairs were randomly selected to perform makeup transfer, resulting in a total of 3,000 transfer results per method for comprehensive evaluation.

\subsubsection{Evaluation Datasets.} We evaluate all methods on BeautyBench and two widely used makeup transfer datasets, MT~\cite{li2018beautygan} and LADN~\cite{gu2019ladn} following prior works. For each dataset, we randomly sample 1,000 source-reference pairs, yielding 3,000 transfer results per method.
%{\bf Evaluation Metrics}. Following previous works, we first employ several standard automatic metrics to evaluate the performance of makeup transfer methods from different perspectives. To assess makeup fidelity, we compute the cosine similarity between the generated and reference images within the CLIP \cite{radford2021learning} embedding space, referred to as the CLIP-I score. Following PhotoMaker\cite{li2024photomaker}, we utilize Face similarity metric to measure the ID consistency, which calculates the cosine similarity of face embedding\cite{schroff2015facenet} between source and transferred results. In addition, we use L2M to measure background preservation over non-facial regions defined by facial parsing\cite{yu2018bisenet}.Since makeup transfer requires both makeup fidelity and identity consistency, achieving high scores in one metric while failing in the other cannot be considered a successful transfer. Therefore, in addition to the aforementioned metrics, we further use the product of CLIP-I and Face Sim, denoted by $C \times F$, for a more comprehensive evaluation. As both CLIP-I and Face Similarity are cosine-based similarity measures, they share the same scale, allowing their product to be directly used as a unified score. A high score can only be achieved when both makeup fidelity and identity consistency are well preserved.

\subsubsection{Evaluation Metrics.} Following prior work, we evaluate makeup fidelity using CLIP-I and identity consistency using Face Similarity. CLIP-I computes cosine similarity between generated and reference images in the CLIP embedding space~\cite{radford2021learning}, while Face Similarity measures cosine similarity between face embeddings of source and transferred images~\cite{schroff2015facenet}. Since successful makeup transfer requires both makeup fidelity and source-identity preservation, neither metric alone provides a complete assessment. We therefore additionally report $C \times F$, the product of CLIP-I and Face Similarity, which yields a high score only when both objectives are well preserved. L2M evaluates background preservation as reconstruction error over non-facial regions obtained via face parsing~\cite{yu2018bisenet}. All metrics are computed per image pair and then averaged over the entire benchmark.

\subsection{Qualitative Analyses}
We compare our method with six representative makeup transfer methods in Figure~\ref{comparison}, including EleGANt \cite{yang2022elegant}, PSGAN \cite{jiang2020psgan}, SSAT \cite{sun2022ssat}, SHMT \cite{sun2024shmt}, Stable-Makeup \cite{zhang2025stablemakeup}, and FLUX-Makeup \cite{zhu2025flux}. As shown in Figure~\ref{comparison}, these methods often fail to faithfully transfer complex makeup patterns, and some methods cause noticeable changes to the source’s facial features or skin tone. For example, Stable-Makeup shifts facial features and expressions toward the reference, whereas SHMT produces facial distortions when the source and reference are significantly different. FLUX-Makeup better preserves source identity but often produces inaccurate complex makeup patterns, especially in their structure and fine details. In contrast, our method faithfully transfers diverse makeup patterns while preserving the source identity, expression, skin tone, and background, producing realistic and natural results. The zoomed-in comparisons in Figure~\ref{details} further demonstrate its advantage. Specifically, our method better captures the decorative elements around the eyes in the first row and generates complex face-painting details accurately in the second.

\begin{table*}[]
\centering
\caption{Quantitative comparison of different makeup transfer methods across three datasets. \colorbox[HTML]{FCE8E7}{\textbf{Red bold}} values indicate the best results, and \colorbox[HTML]{ECF4FF}{\underline{blue underlined}} values indicate the second-best results. $C$ and $F$ refer to the CLIP-I score and Face Similarity.  $C \times F$ denotes the product of these two metrics.}
\label{tab:comparison}
\footnotesize
\resizebox{\textwidth}{!}{%
\renewcommand\arraystretch{1.3}
\begin{tabular}{@{}l|cccc|cccc|cccc@{}}
\toprule
& \multicolumn{4}{c|}{\textbf{MT}}
& \multicolumn{4}{c|}{\textbf{LADN}}
& \multicolumn{4}{c}{\textbf{BeautyBench}}
\\
\midrule

\textbf{Method}
& \textbf{L2M$\downarrow$}
& \textbf{C$\uparrow$}
& \textbf{F$\uparrow$}
& \textbf{C$\times$F $\uparrow$}
& \textbf{L2M$\downarrow$}
& \textbf{C$\uparrow$}
& \textbf{F$\uparrow$}
& \textbf{C$\times$F $\uparrow$}
& \textbf{L2M$\downarrow$}
& \textbf{C$\uparrow$}
& \textbf{F$\uparrow$}
& \textbf{C$\times$F $\uparrow$}
\\
\midrule

\textbf{CPM}
& 0.077 & 0.618 & 0.621 & 0.386  % MT
& 0.136 & 0.647 & 0.589 & 0.382  % LADN
& 0.177 & 0.503 & 0.509 & 0.252  % BeautyBench
\\

\textbf{EleGANt}
& 0.117 & 0.603 & 0.861 & 0.521  % MT
& 0.067 & 0.582 & 0.839 & 0.490  % LADN
& 0.090 & 0.440
& \colorbox[HTML]{ECF4FF}{\underline{0.833}}
& 0.367                              % BeautyBench
\\

\textbf{PSGAN}
& 0.114 & 0.596 & 0.836 & 0.499  % MT
& 0.061 & 0.565 & 0.830 & 0.470  % LADN
& 0.082 & 0.431 & 0.824 & 0.355  % BeautyBench
\\

\textbf{SSAT}
& 0.183 & 0.592 & 0.788 & 0.468  % MT
& 0.218 & 0.609 & 0.769 & 0.471  % LADN
& 0.190 & 0.471 & 0.731 & 0.344  % BeautyBench
\\

\textbf{SHMT}
& 0.060
& \colorbox[HTML]{ECF4FF}{\underline{0.643}}
& 0.493
& 0.319                              % MT
& 0.060 & 0.702 & 0.416 & 0.293  % LADN
& 0.066 & 0.508 & 0.363 & 0.184  % BeautyBench
\\

\textbf{Stable-Makeup}
& 0.046
& \colorbox[HTML]{FCE8E7}{\textbf{0.708}}
& 0.579
& 0.411                              % MT
& \colorbox[HTML]{ECF4FF}{\underline{0.048}}
& \colorbox[HTML]{FCE8E7}{\textbf{0.773}}
& 0.528
& 0.410                              % LADN
& 0.054
& \colorbox[HTML]{FCE8E7}{\textbf{0.656}}
& 0.374
& 0.246                              % BeautyBench
\\

\textbf{FLUX-Makeup}
& \colorbox[HTML]{ECF4FF}{\underline{0.033}}
& 0.632
& \colorbox[HTML]{ECF4FF}{\underline{0.901}}
& \colorbox[HTML]{ECF4FF}{\underline{0.570}}   % MT
& \colorbox[HTML]{FCE8E7}{\textbf{0.035}}
& 0.695
& \colorbox[HTML]{ECF4FF}{\underline{0.859}}
& \colorbox[HTML]{ECF4FF}{\underline{0.598}}   % LADN
& \colorbox[HTML]{ECF4FF}{\underline{0.040}}
& 0.531
& 0.784
& \colorbox[HTML]{ECF4FF}{\underline{0.414}}   % BeautyBench
\\

\midrule

\textbf{Ours}
& \colorbox[HTML]{FCE8E7}{\textbf{0.029}}
& \colorbox[HTML]{ECF4FF}{\underline{0.643}}
& \colorbox[HTML]{FCE8E7}{\textbf{0.954}}
& \colorbox[HTML]{FCE8E7}{\textbf{0.612}}   % MT
& 0.063
& \colorbox[HTML]{ECF4FF}{\underline{0.703}}
& \colorbox[HTML]{FCE8E7}{{\textbf{0.900}}}
& \colorbox[HTML]{FCE8E7}{\textbf{0.630}}   % LADN
& \colorbox[HTML]{FCE8E7}{\textbf{0.033}}
& \colorbox[HTML]{ECF4FF}{\underline{0.536}}
& \colorbox[HTML]{FCE8E7}{\textbf{0.911}}
& \colorbox[HTML]{FCE8E7}{\textbf{0.484}}   % BeautyBench
\\

\bottomrule
\end{tabular}
}
\end{table*}

% \begin{figure}[h]
% \centering
% \includegraphics[width=8.5cm]{images/experiment.jpg}
% \caption{Visualization of different makeup transfer methods in the two-dimensional space of min-max normalized CLIP-I and Face Similarity. For each metric, min-max normalization is applied across all methods to linearly map scores into $[0,1]$ while preserving their relative positions. Methods with high values on both metrics occupy a larger rectangular area in this space, indicating stronger joint performance in makeup fidelity and identity consistency.}
% \label{experiment}
% \end{figure}

\subsection{Quantitative Evaluations}

We evaluate makeup transfer performance using four metrics: L2M, CLIP-I, Face Similarity, and $C \times F$. As shown in Table~\ref{tab:comparison}, our method achieves the highest $C \times F$ and Face Similarity across all three benchmarks, while obtaining the best L2M on MT and BeautyBench. In addition, it achieves the second-best CLIP-I across all three benchmarks. Although Stable-Makeup achieves the highest CLIP-I, its lower Face Similarity suggests that the high score may partly arise from identity drift toward the reference image rather than more faithful makeup transfer. This tendency is also evident in the qualitative comparisons. Conversely, EleGANt, PSGAN, and SSAT achieve relatively high Face Similarity but comparatively low CLIP-I scores, indicating insufficient makeup transfer. FLUX-Makeup consistently ranks second in $C \times F$ across all three benchmarks, indicating a relatively balanced performance between makeup fidelity and identity preservation. However, our method consistently surpasses FLUX-Makeup in both CLIP-I and Face Similarity, with a particularly clear advantage on the challenging BeautyBench. Overall, our method achieves a stronger balance between makeup fidelity and identity consistency, maintaining high makeup similarity while providing particularly strong identity preservation.

\begin{figure}[t]
        \centering
        \includegraphics[width=7.5cm]{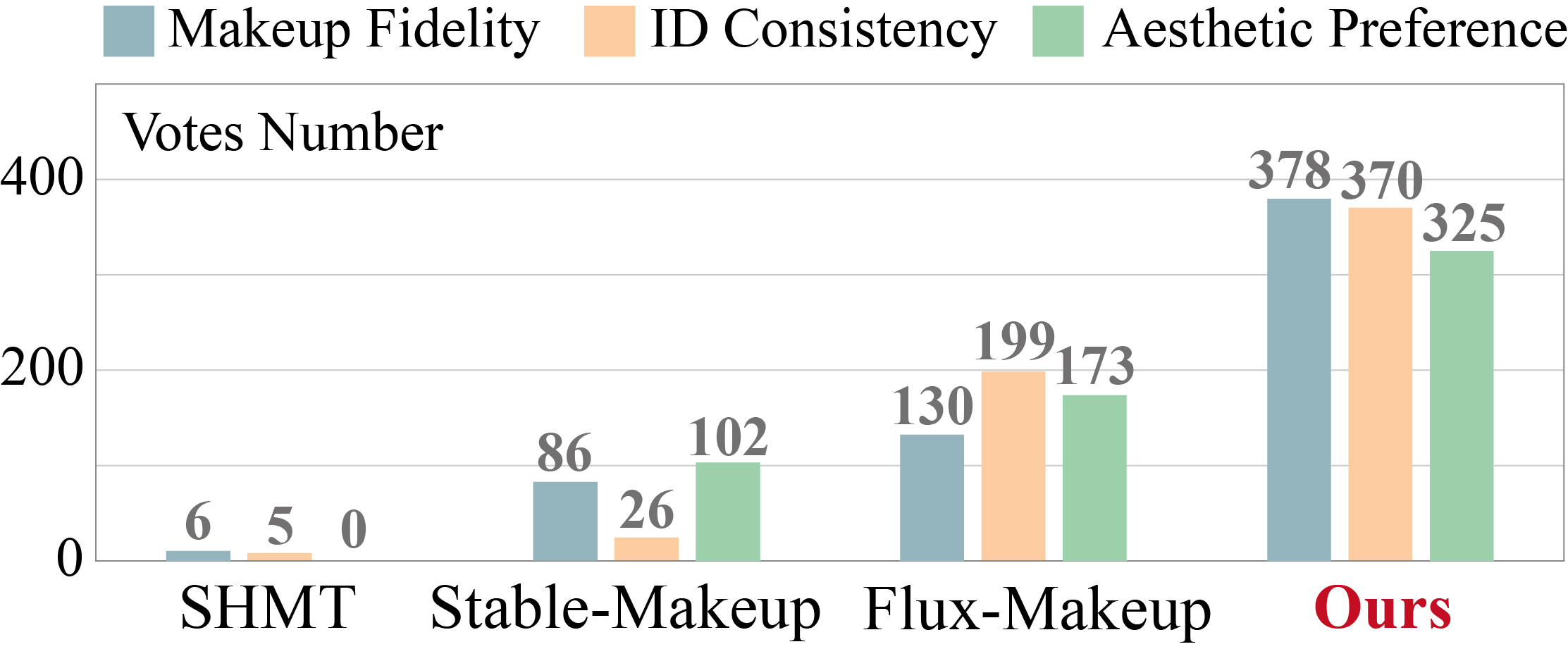}
        \caption{User study results comparing our method with SHMT, Stable-Makeup, and FLUX-Makeup in makeup fidelity, identity consistency, and aesthetic preference. }
        \label{user_study}
        \vspace{-2mm}
\end{figure}

% \caption{Ablation study of the reinforcement learning stage on MT, LADN, and BeautyBench. L2M denotes background preservation, while $C$ and $F$ denote the CLIP-I score and Face Similarity, respectively. $C \times F$ represents the product of these two metrics. The results indicate that the RL stage maintains performance comparable to the SFT-only baseline across all three benchmarks.}

\begin{table*}[t]
\centering
\caption{Ablation study of the reinforcement learning stage on BeautyBench and its three makeup categories. $C$ and $F$ denote the CLIP-I score and Face Similarity, respectively, while $C \times F$ represents their product.}
\label{tab:ablation}
\footnotesize
\renewcommand{\arraystretch}{1.2}
\begin{tabularx}{\textwidth}{@{}l|YYY|YYY|YYY|YYY@{}}
\toprule
& \multicolumn{3}{c|}{\textbf{BeautyBench}}
& \multicolumn{3}{c|}{\textbf{Light Makeup}}
& \multicolumn{3}{c|}{\textbf{Heavy Makeup}}
& \multicolumn{3}{c}{\textbf{Complex Makeup}} \\ 
\midrule
\textbf{Method}
& \textbf{C$\uparrow$} 
& \textbf{F$\uparrow$} 
& \multicolumn{1}{c|}{\textbf{C$\times$F$\uparrow$}}
& \textbf{C$\uparrow$} 
& \textbf{F$\uparrow$} 
& \multicolumn{1}{c|}{\textbf{C$\times$F$\uparrow$}}
& \textbf{C$\uparrow$} 
& \textbf{F$\uparrow$} 
& \multicolumn{1}{c|}{\textbf{C$\times$F$\uparrow$}}
& \textbf{C$\uparrow$} 
& \textbf{F$\uparrow$} 
& \textbf{C$\times$F$\uparrow$} \\ 
\midrule
\textbf{SFT}
& 0.533          & 0.893          & 0.472
& 0.502          & 0.954          & 0.480
& 0.511          & 0.897          & 0.458
& 0.627          & 0.775          & 0.481 \\

\textbf{SFT+RL}
& \textbf{0.536} & \textbf{0.911} & \textbf{0.484}
& \textbf{0.505} & \textbf{0.969} & \textbf{0.489}
& \textbf{0.514} & \textbf{0.921} & \textbf{0.473}
& \textbf{0.633} & \textbf{0.787} & \textbf{0.493} \\
\bottomrule
\end{tabularx}
\end{table*}

\subsection{User Study}
In the user study, we compared our method with three representative makeup transfer approaches: SHMT \cite{sun2024shmt}, Stable-Makeup \cite{zhang2025stablemakeup}, and FLUX-Makeup \cite{zhu2025flux}. For each makeup transfer sample, participants were shown the results produced by the four methods and asked to select the best result in terms of makeup fidelity, identity consistency, and aesthetic preference, respectively. The detailed questions are provided in the supplementary material. To ensure objectivity, the method names were concealed, and the order of images was randomized. A total of 30 participants assessed four methods on 20 randomly selected samples from BeautyBench, yielding 1,800 valid votes. Before participation, all participants were informed about the purpose and procedure of the user study, and their informed consent was obtained. As shown in Figure~\ref{user_study}, our method received the most votes across all three criteria, substantially outperforming the competing methods.

\subsection{Ablation Study}
\begin{figure}[t]
        \centering
        \includegraphics[width=7.8cm]{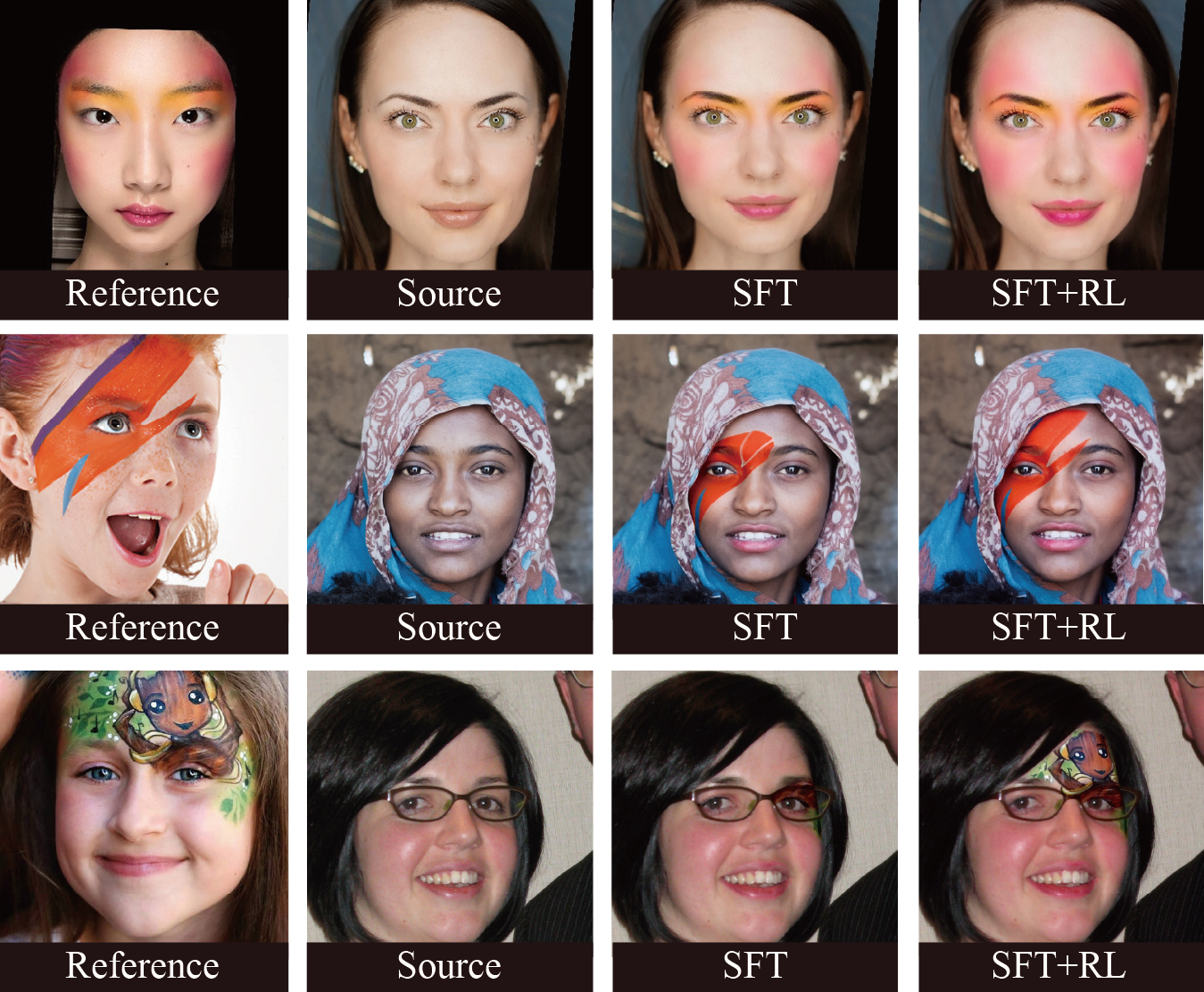}
        \caption{Qualitative comparison between the SFT-only and SFT+RL models. The RL model mitigates the insufficient transfer issue of the SFT-only model, leading to improved visual quality in makeup transfer.}
        \label{ablation_image}
        \vspace{-3mm}
\end{figure}

Table~\ref{tab:ablation} compares the fully trained model (SFT + RL) with the SFT-only model on the BeautyBench. It further reports their performance on three category-specific subsets: light, heavy, and complex makeup. The results show that the fully trained model consistently outperforms the SFT-only model on the full benchmark and all three subsets. Face Similarity improves consistently across all three makeup categories, while the gain in CLIP-I is most evident for complex makeup. Moreover, RL training yields qualitative improvements, as shown in Figure~\ref{ablation_image}. In the first row, the fully trained model produces a more pronounced makeup effect, demonstrating better fidelity in the transfer of makeup details. The second and third rows present more complex cases. RL training helps reduce insufficient or failed transfer. This demonstrates the effectiveness of the proposed framework in further enhancing the model's ability to handle complex scenarios.

\subsection{Benchmark Analysis for Makeup Transfer}
Table \ref{tab:benchmark} reports the statistics of MT, LADN, and BeautyBench with respect to skin tone, gender, and makeup style. In terms of skin-tone distribution, BeautyBench shows the highest distribution entropy of 1.951 bits. Regarding gender distribution, it presents a balanced split, whereas the other two benchmarks are dominated by female subjects. As for makeup style, we categorize the makeup images into two types: normal makeup and complex makeup. Complex makeup includes face painting and facial decorations, while normal makeup refers to more conventional styles without such fancy designs. Among the three benchmarks, BeautyBench contains the largest proportion of complex makeup samples. These statistics show that BeautyBench provides a more diverse and challenging benchmark for evaluating makeup transfer methods. 
Additionally, as illustrated in Table \ref{tab:comparison}, existing methods achieve noticeably lower CLIP-I and $C \times F$ on BeautyBench than on MT and LADN. This further demonstrates that BeautyBench provides a more demanding evaluation that better reveals the limitations of existing methods.

\begin{table}[t]
\centering
\caption{Benchmark statistics of MT, LADN, and BeautyBench, showcasing the diversity and comprehensiveness of our benchmark.}
\label{tab:benchmark}
\footnotesize
\renewcommand{\arraystretch}{1.1}

\resizebox{\columnwidth}{!}{%
\begin{tabular}{@{}c|c|ccc@{}}
\toprule

\multicolumn{2}{c|}{\textbf{Benchmarks}}
& \textbf{MT}
& \textbf{LADN}
& \makecell[c]{\textbf{Beauty}\\\textbf{Bench}}
\\
\midrule

\makecell[c]{\textbf{Skin}\\\textbf{Tone}}
& \makecell[c]{Distribution\\Entropy$\uparrow$}
& 1.185 bits
& 1.449 bits
& \textbf{1.951} bits
\\
\midrule

\multirow[c]{2}{*}{\textbf{Gender}}
& Male
& 22.4\%
& 11.7\%
& 50.0\%
\\

& Female
& 77.6\%
& 88.3\%
& 50.0\%
\\
\midrule

\multirow[c]{2}{*}{\makecell[c]{\textbf{Makeup}\\\textbf{Style}}}
& Normal
& 98.5\%
& 82.0\%
& 76.7\%
\\

& Complex
& 1.5\%
& 18.0\%
& 23.3\%
\\
\bottomrule
\end{tabular}%
}
\vspace{-2mm}
\end{table}

\section{Conclusion}
In this work, we propose a novel data curation pipeline for producing high-quality supervised training data and construct a new dataset, ConsistentBeauty. We further introduce RealBeauty, a synthetic-to-real post-training framework for makeup transfer, enabling models to learn effectively from both synthetic and real data. In addition, we establish a more challenging benchmark that provides a more comprehensive evaluation of this task. Extensive experiments provide consistent evidence for the effectiveness of our approach.

\bibliography{refs}

@String(ECCV= {Eur. Conf. Comput. Vis.})

@String(AAAI = {AAAI})

@String(ECCV  = {ECCV})

@inproceedings{li2018beautygan,
  title={Beautygan: Instance-level facial makeup transfer with deep generative adversarial network},
  author={Li, Tingting and Qian, Ruihe and Dong, Chao and Liu, Si and Yan, Qiong and Zhu, Wenwu and Lin, Liang},
  booktitle={Proceedings of the 26th ACM international conference on Multimedia},
  pages={645--653},
  year={2018}
}

@inproceedings{jiang2020psgan,
  title={Psgan: Pose and expression robust spatial-aware gan for customizable makeup transfer},
  author={Jiang, Wentao and Liu, Si and Gao, Chen and Cao, Jie and He, Ran and Feng, Jiashi and Yan, Shuicheng},
  booktitle={Proceedings of the IEEE/CVF conference on computer vision and pattern recognition},
  pages={5194--5202},
  year={2020}
}

@inproceedings{yang2022elegant,
  title={Elegant: Exquisite and locally editable gan for makeup transfer},
  author={Yang, Chenyu and He, Wanrong and Xu, Yingqing and Gao, Yang},
  booktitle={European conference on computer vision},
  pages={737--754},
  year={2022},
  organization={Springer}
}

@inproceedings{sun2022ssat,
  title={Ssat: A symmetric semantic-aware transformer network for makeup transfer and removal},
  author={Sun, Zhaoyang and Chen, Yaxiong and Xiong, Shengwu},
  booktitle={Proceedings of the AAAI Conference on artificial intelligence},
  volume={36},
  number={2},
  pages={2325--2334},
  year={2022}
}

@inproceedings{zhang2025stablemakeup,
  title={Stablemakeup: When real-world makeup transfer meets diffusion model},
  author={Zhang, Yuxuan and Yuan, Yirui and Song, Yiren and Liu, Jiaming},
  booktitle={Proceedings of the Special Interest Group on Computer Graphics and Interactive Techniques Conference Conference Papers},
  pages={1--9},
  year={2025}
}

@article{yang2025ffhq,
  title={FFHQ-Makeup: Paired Synthetic Makeup Dataset with Facial Consistency Across Multiple Styles},
  author={Yang, Xingchao and Ueda, Shiori and Huang, Yuantian and Akiyama, Tomoya and Taketomi, Takafumi},
  journal={arXiv preprint arXiv:2508.03241},
  year={2025}
}

@article{zhu2025flux,
  title={FLUX-Makeup: High-Fidelity, Identity-Consistent, and Robust Makeup Transfer via Diffusion Transformer},
  author={Zhu, Jian and Liu, Shanyuan and Li, Liuzhuozheng and Gong, Yue and Wang, He and Cheng, Bo and Ma, Yuhang and Wu, Liebucha and Wu, Xiaoyu and Leng, Dawei and others},
  journal={arXiv preprint arXiv:2508.05069},
  year={2025}
}

@article{wu2025evomakeup,
  title={EvoMakeup: High-Fidelity and Controllable Makeup Editing with MakeupQuad},
  author={Wu, Huadong and Fu, Yi and Li, Yunhao and Gao, Yuan and Du, Kang},
  journal={arXiv preprint arXiv:2508.05994},
  year={2025}
}

@inproceedings{gu2019ladn,
  title={Ladn: Local adversarial disentangling network for facial makeup and de-makeup},
  author={Gu, Qiao and Wang, Guanzhi and Chiu, Mang Tik and Tai, Yu-Wing and Tang, Chi-Keung},
  booktitle={Proceedings of the IEEE/CVF International conference on computer vision},
  pages={10481--10490},
  year={2019}
}

@inproceedings{yan2023beautyrec,
  title={Beautyrec: Robust, efficient, and component-specific makeup transfer},
  author={Yan, Qixin and Guo, Chunle and Zhao, Jixin and Dai, Yuekun and Loy, Chen Change and Li, Chongyi},
  booktitle={Proceedings of the IEEE/CVF conference on computer vision and pattern recognition},
  pages={1102--1110},
  year={2023}
}

@article{goodfellow2020generative,
  title={Generative adversarial networks},
  author={Goodfellow, Ian and Pouget-Abadie, Jean and Mirza, Mehdi and Xu, Bing and Warde-Farley, David and Ozair, Sherjil and Courville, Aaron and Bengio, Yoshua},
  journal={Communications of the ACM},
  volume={63},
  number={11},
  pages={139--144},
  year={2020},
  publisher={ACM New York, NY, USA}
}

@article{sun2024shmt,
  title={Shmt: Self-supervised hierarchical makeup transfer via latent diffusion models},
  author={Sun, Zhaoyang and Xiong, Shengwu and Chen, Yaxiong and Du, Fei and Chen, Weihua and Wang, Fan and Rong, Yi},
  journal={Advances in Neural Information Processing Systems},
  volume={37},
  pages={16016--16042},
  year={2024}
}

@article{labs2025flux,
  title={FLUX. 1 Kontext: Flow Matching for In-Context Image Generation and Editing in Latent Space},
  author={Labs, Black Forest and Batifol, Stephen and Blattmann, Andreas and Boesel, Frederic and Consul, Saksham and Diagne, Cyril and Dockhorn, Tim and English, Jack and English, Zion and Esser, Patrick and others},
  journal={arXiv preprint arXiv:2506.15742},
  year={2025}
}

@article{wang2025seededit,
  title={SeedEdit 3.0: Fast and High-Quality Generative Image Editing},
  author={Wang, Peng and Shi, Yichun and Lian, Xiaochen and Zhai, Zhonghua and Xia, Xin and Xiao, Xuefeng and Huang, Weilin and Yang, Jianchao},
  journal={arXiv preprint arXiv:2506.05083},
  year={2025}
}

@article{ho2020denoising,
  title={Denoising diffusion probabilistic models},
  author={Ho, Jonathan and Jain, Ajay and Abbeel, Pieter},
  journal={Advances in neural information processing systems},
  volume={33},
  pages={6840--6851},
  year={2020}
}

@article{song2019generative,
  title={Generative modeling by estimating gradients of the data distribution},
  author={Song, Yang and Ermon, Stefano},
  journal={Advances in neural information processing systems},
  volume={32},
  year={2019}
}

@inproceedings{rombach2022high,
  title={High-resolution image synthesis with latent diffusion models},
  author={Rombach, Robin and Blattmann, Andreas and Lorenz, Dominik and Esser, Patrick and Ommer, Bj{\"o}rn},
  booktitle={Proceedings of the IEEE/CVF conference on computer vision and pattern recognition},
  pages={10684--10695},
  year={2022}
}

@inproceedings{esser2024scaling,
  title={Scaling rectified flow transformers for high-resolution image synthesis},
  author={Esser, Patrick and Kulal, Sumith and Blattmann, Andreas and Entezari, Rahim and M{\"u}ller, Jonas and Saini, Harry and Levi, Yam and Lorenz, Dominik and Sauer, Axel and Boesel, Frederic and others},
  booktitle={Forty-first international conference on machine learning},
  year={2024}
}

@article{liu2022flow,
  title={Flow straight and fast: Learning to generate and transfer data with rectified flow},
  author={Liu, Xingchao and Gong, Chengyue and Liu, Qiang},
  journal={arXiv preprint arXiv:2209.03003},
  year={2022}
}

@article{lipman2022flow,
  title={Flow matching for generative modeling},
  author={Lipman, Yaron and Chen, Ricky TQ and Ben-Hamu, Heli and Nickel, Maximilian and Le, Matt},
  journal={arXiv preprint arXiv:2210.02747},
  year={2022}
}

@article{lee2023aligning,
  title={Aligning text-to-image models using human feedback},
  author={Lee, Kimin and Liu, Hao and Ryu, Moonkyung and Watkins, Olivia and Du, Yuqing and Boutilier, Craig and Abbeel, Pieter and Ghavamzadeh, Mohammad and Gu, Shixiang Shane},
  journal={arXiv preprint arXiv:2302.12192},
  year={2023}
}

@inproceedings{wallace2024diffusion,
  title={Diffusion model alignment using direct preference optimization},
  author={Wallace, Bram and Dang, Meihua and Rafailov, Rafael and Zhou, Linqi and Lou, Aaron and Purushwalkam, Senthil and Ermon, Stefano and Xiong, Caiming and Joty, Shafiq and Naik, Nikhil},
  booktitle={Proceedings of the IEEE/CVF Conference on Computer Vision and Pattern Recognition},
  pages={8228--8238},
  year={2024}
}

@article{rafailov2023direct,
  title={Direct preference optimization: Your language model is secretly a reward model},
  author={Rafailov, Rafael and Sharma, Archit and Mitchell, Eric and Manning, Christopher D and Ermon, Stefano and Finn, Chelsea},
  journal={Advances in neural information processing systems},
  volume={36},
  pages={53728--53741},
  year={2023}
}

@article{shao2024deepseekmath,
  title={Deepseekmath: Pushing the limits of mathematical reasoning in open language models},
  author={Shao, Zhihong and Wang, Peiyi and Zhu, Qihao and Xu, Runxin and Song, Junxiao and Bi, Xiao and Zhang, Haowei and Zhang, Mingchuan and Li, YK and Wu, Yang and others},
  journal={arXiv preprint arXiv:2402.03300},
  year={2024}
}

@article{liu2025flow,
  title={Flow-grpo: Training flow matching models via online rl},
  author={Liu, Jie and Liu, Gongye and Liang, Jiajun and Li, Yangguang and Liu, Jiaheng and Wang, Xintao and Wan, Pengfei and Zhang, Di and Ouyang, Wanli},
  journal={arXiv preprint arXiv:2505.05470},
  year={2025}
}

@article{xue2025dancegrpo,
  title={DanceGRPO: Unleashing GRPO on Visual Generation},
  author={Xue, Zeyue and Wu, Jie and Gao, Yu and Kong, Fangyuan and Zhu, Lingting and Chen, Mengzhao and Liu, Zhiheng and Liu, Wei and Guo, Qiushan and Huang, Weilin and others},
  journal={arXiv preprint arXiv:2505.07818},
  year={2025}
}

@article{li2025mixgrpo,
  title={Mixgrpo: Unlocking flow-based grpo efficiency with mixed ode-sde},
  author={Li, Junzhe and Cui, Yutao and Huang, Tao and Ma, Yinping and Fan, Chun and Yang, Miles and Zhong, Zhao},
  journal={arXiv preprint arXiv:2507.21802},
  year={2025}
}

@inproceedings{caron2021emerging,
  title={Emerging properties in self-supervised vision transformers},
  author={Caron, Mathilde and Touvron, Hugo and Misra, Ishan and J{\'e}gou, Herv{\'e} and Mairal, Julien and Bojanowski, Piotr and Joulin, Armand},
  booktitle={Proceedings of the IEEE/CVF international conference on computer vision},
  pages={9650--9660},
  year={2021}
}

@inproceedings{karras2019style,
  title={A style-based generator architecture for generative adversarial networks},
  author={Karras, Tero and Laine, Samuli and Aila, Timo},
  booktitle={Proceedings of the IEEE/CVF conference on computer vision and pattern recognition},
  pages={4401--4410},
  year={2019}
}

@inproceedings{nguyen2021lipstick,
  title={Lipstick ain't enough: Beyond color matching for in-the-wild makeup transfer},
  author={Nguyen, Thao and Tran, Anh Tuan and Hoai, Minh},
  booktitle={Proceedings of the IEEE/CVF Conference on computer vision and pattern recognition},
  pages={13305--13314},
  year={2021}
}

@inproceedings{schroff2015facenet,
  title={Facenet: A unified embedding for face recognition and clustering},
  author={Schroff, Florian and Kalenichenko, Dmitry and Philbin, James},
  booktitle={Proceedings of the IEEE conference on computer vision and pattern recognition},
  pages={815--823},
  year={2015}
}

@inproceedings{yu2018bisenet,
  title={Bisenet: Bilateral segmentation network for real-time semantic segmentation},
  author={Yu, Changqian and Wang, Jingbo and Peng, Chao and Gao, Changxin and Yu, Gang and Sang, Nong},
  booktitle={Proceedings of the European conference on computer vision (ECCV)},
  pages={325--341},
  year={2018}
}

@inproceedings{radford2021learning,
  title={Learning transferable visual models from natural language supervision},
  author={Radford, Alec and Kim, Jong Wook and Hallacy, Chris and Ramesh, Aditya and Goh, Gabriel and Agarwal, Sandhini and Sastry, Girish and Askell, Amanda and Mishkin, Pamela and Clark, Jack and others},
  booktitle={International conference on machine learning},
  pages={8748--8763},
  year={2021},
  organization={PmLR}
}

@inproceedings{zhu2017unpaired,
  title={Unpaired image-to-image translation using cycle-consistent adversarial networks},
  author={Zhu, Jun-Yan and Park, Taesung and Isola, Phillip and Efros, Alexei A},
  booktitle={Proceedings of the IEEE international conference on computer vision},
  pages={2223--2232},
  year={2017}
}

@article{song2020denoising,
  title={Denoising diffusion implicit models},
  author={Song, Jiaming and Meng, Chenlin and Ermon, Stefano},
  journal={arXiv preprint arXiv:2010.02502},
  year={2020}
}

@article{song2020score,
  title={Score-based generative modeling through stochastic differential equations},
  author={Song, Yang and Sohl-Dickstein, Jascha and Kingma, Diederik P and Kumar, Abhishek and Ermon, Stefano and Poole, Ben},
  journal={arXiv preprint arXiv:2011.13456},
  year={2020}
}

@article{ramesh2022hierarchical,
  title={Hierarchical text-conditional image generation with clip latents},
  author={Ramesh, Aditya and Dhariwal, Prafulla and Nichol, Alex and Chu, Casey and Chen, Mark},
  journal={arXiv preprint arXiv:2204.06125},
  volume={1},
  number={2},
  pages={3},
  year={2022}
}

@inproceedings{huang2026competition,
  title={From Competition to Synergy: Unlocking Reinforcement Learning for Subject-Driven Image Generation},
  author={Huang, Ziwei and Shu, Ying and Long, Quanyu and Wang, Wenya and Guo, Qiushi and Ge, Tiezheng and Gan, Leilei and others},
  booktitle={Proceedings of the 64th Annual Meeting of the Association for Computational Linguistics (Volume 1: Long Papers)},
  pages={41117--41136},
  year={2026}
}

\vfill

% \bibliography{aaai2027}


\end{document}